\documentclass{article}
\usepackage{spconf,amsmath,graphicx}

\usepackage{xspace}
\usepackage{array}
\usepackage{eqparbox}
\usepackage{url}
\usepackage{longtable}
\usepackage{lipsum}
\usepackage{blindtext}
\usepackage{makecell}
\usepackage{mathtools}
\usepackage{commath}
\usepackage{multirow}
\usepackage{tabularx}
\usepackage[normalem]{ulem}
\usepackage{xspace}
\usepackage{algorithm}
\usepackage{algpseudocode}
\usepackage{amsfonts}
\usepackage{placeins}
\usepackage{booktabs}
\usepackage[normalem]{ulem}

\newcommand{\ind}{\mathbb{I}}
\newcolumntype{C}{>{\hsize=\dimexpr0.5\hsize+8\tabcolsep+\arrayrulewidth\centering\relax}X}

\DeclareMathOperator*{\argmax}{arg\,max}
\DeclareMathOperator{\cref}{ref}

\DeclareMathOperator{\softmax}{softmax}
\DeclareMathOperator{\shuffle}{shuffle}
\DeclareMathOperator{\sharpen}{sharpen}

\title{Holistic Semi-Supervised Approaches for EEG Representation Learning}

\name{Guangyi Zhang and Ali Etemad
}
\address{Department of Electrical and Computer Engineering\\
Queen's University, Kingston, Canada}
\begin{document}
% \ninept
\maketitle

\begin{abstract}
Recently, supervised methods, which often require substantial amounts of class labels, have achieved promising results for EEG representation learning. However, labeling EEG data is a challenging task. More recently, holistic semi-supervised learning approaches, which only require few output labels, have shown promising results in the field of computer vision. These methods, however, have not yet been adapted for EEG learning. In this paper, we adapt three state-of-the-art holistic semi-supervised approaches, namely MixMatch \cite{berthelot2019mixmatch}, FixMatch \cite{sohn2020fixmatch}, and AdaMatch \cite{berthelot2021adamatch}, as well as five classical semi-supervised methods for EEG learning. We perform rigorous experiments with all $8$ methods on two public EEG-based emotion recognition datasets, namely SEED and SEED-IV. The experiments with different amounts of limited labeled samples show that the holistic approaches achieve strong results even when only 1 labeled sample is used per class. Further experiments show that in most cases, AdaMatch is the most effective method, followed by MixMatch and FixMatch. 

\end{abstract}

\section{Introduction} \label{sec:intro}
Human emotions are highly informative non-verbal cues that can be widely used to enhance human-machine interaction. Many solutions have been proposed for affective computing \cite{picard2000affective} using Electroencephalography (EEG), as EEG is widely used for directly measuring brain activity with high spatio-temporal resolution. Recently, various deep learning models have achieved state-of-the-art performances in emotion recognition tasks due to the capability of learning highly discriminative information from multichannel EEG recordings \cite{zheng2015investigating,zheng2018emotionmeter}. However, as the significant majority of existing EEG-based deep emotion recognition solutions are `\textit{supervised}' methods \cite{zheng2018emotionmeter,zhang2021capsule,zhang2020rfnet, zheng2017identifying,zhang2018spatial}, they rely on the annotated labels.
% Labeling EEG, on the other hand, is difficult, time-consuming, costly, and often requires professional annotators \cite{zheng2015investigating,zheng2018emotionmeter}. 
Labeling EEG, on the other hand, is difficult, time-consuming, and costly. To obtain reliable labels, many EEG-based studies require various emotion annotation approaches, including pre-stimulation and post-experiment self-assessments, as well as multiple expert evaluations \cite{zheng2015investigating,zheng2018emotionmeter,correa2018amigos}. 
Moreover, most existing fully supervised learning algorithms suffer from performance drop when only a few training samples are labeled.
As a result, it is important to investigate how to develop effective solutions that can deal with limited labeled training samples.

Semi-Supervised Learning (SSL) has been recognized as a powerful paradigm to address the problems of a scarcity of labeled training samples, achieving great success in the field of computer vision \cite{van2020survey}. For example, pseudo-labeling was proposed to encourage low-entropy predictions of unlabeled data \cite{lee2013pseudo}. To do so, a model was first trained using limited labeled data. The model was then used to predict pseudo-labels for the unlabeled data. Finally, the model was retrained using the entire data with true labels and pseudo labels combined \cite{lee2013pseudo}. Consistency regularization, in conjunction with data augmentation, has lately become popular in SSL studies \cite{samuli2017temporal,tarvainen2017mean,oliver2018realistic}. Common regularization techniques including stochastic augmentation applied to input data and dropout applied throughout the network have been employed in recent SSL frameworks \cite{samuli2017temporal}. For instance, the $\Pi$-model \cite{samuli2017temporal} trained the network (with dropout) on both the original and augmented inputs, and minimized the distance between their corresponding outputs, as an unsupervised loss. Meanwhile, a supervised cross-entropy loss was only computed for the labeled set. To improve the $\Pi$-model, `temporal ensembling' and `mean teacher' methods further relied on consistency regularization techniques for SSL \cite{samuli2017temporal,tarvainen2017mean}. 

\begin{figure}[t!] 
    \begin{center}
    \includegraphics[width=0.8\columnwidth]{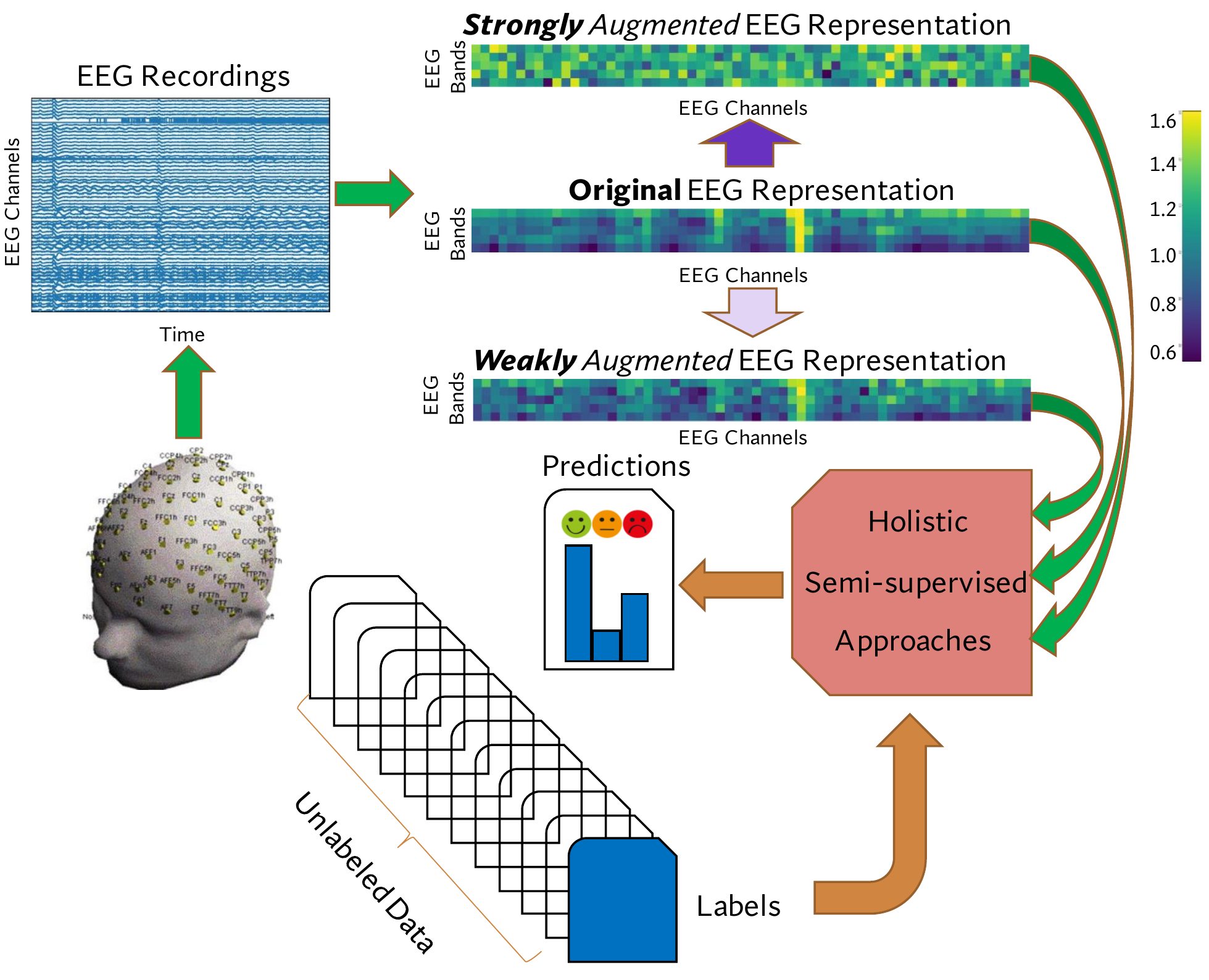} 
    \end{center}
\caption{An overview of holistic SSL approaches for EEG representation learning in the training phase is presented.}
\vspace{-3mm}
\label{fig: overview}
\end{figure}

In addition to the classical SSL approaches mentioned above, several new holistic SSL methods \cite{berthelot2019mixmatch,sohn2020fixmatch,berthelot2021adamatch} have been recently proposed and achieved state-of-the-art results (we discuss these in detail in the next Section). However, their use has been mostly limited to computer vision tasks, and have therefore not been explored in other domains such as brain-computer interfaces (BCI) or biological signal processing. Only in a recent paper \cite{zhang2021deep}, few of the classical SSL frameworks were used for BCI with EEG representation learning. In this paper, we \textbf{\textit{adapt, implement, and where necessary, modify}} three state-of-the-art holistic SSL methods, MixMatch \cite{berthelot2019mixmatch}, FixMatch \cite{sohn2020fixmatch}, and AdaMatch \cite{berthelot2021adamatch}, in addition to the five classical SSL methods, $\Pi$-model \cite{samuli2017temporal},temporal assembling \cite{samuli2017temporal}, mean teacher \cite{tarvainen2017mean}, convolutional autoencoders \cite{tong2019caesnet}, and pseudo labeling \cite{lee2013pseudo}, for EEG representation learning. We then \textbf{\textit{perform a comprehensive study}}, comparing these methods against one another when different amounts of labels are used per class. Our study shows that these methods can indeed be effectively used in the field of BCI for EEG learning, with the holistic methods, particularly AdaMatch and MixMatch, showing reasonable performances even when 1 label per class is used for training. An overview of the holistic SSL methods for EEG learning is shown in Figure \ref{fig: overview}.

\section{Holistic Self-supervised EEG Learning}

\subsection{Problem Setup}
Let's denote $\mathcal{D}_l=\{(x_i^l, y_i^l)\}_{i=1}^M$, $\mathcal{D}_u= \{x_i^u\}_{i=1}^N$, $\mathcal{D}_v=\{(x_i^l, y_i^l)\}_{i=1}^I$, and $\mathcal{D}$ as the labeled, unlabeled, validation, and entire training sets, where $x$ and $y$ represent EEG data and class labels. Here, $M$, $N$, and $I$ are the size of the three subsets.
We have $\mathcal{D}_l \cup \mathcal{D}_u \cup \mathcal{D}_v  = \mathcal{D}$ and $\mathcal{D}_l \cap \mathcal{D}_u \cap \mathcal{D}_v= \emptyset$. We randomly choose $M=m\times k$ samples to form $\mathcal{D}_l$, where $k$ and $m$ are the number of output classes and labeled training samples per class. Given our interest in relying only a few labeled samples, we set $M\ll N$.

\noindent \textbf{Notation:} We denote $\{(x_b^l, y_b^l)\}$ and $\{x_b^u\}$ as labeled and unlabeled sets in each training batch, where $B$ is the batch size. Strong and weak augmentations are represented by $\mathcal{A}_s$ and $\mathcal{A}_w$, respectively (discussed in Section 3.4). We also use the terms $p_m$ and $\theta$ to refer to the deep learning model and model parameters, respectively.

\subsection{Holistic SSL Approaches}

\noindent \textbf{MixMatch.}
To adapt MixMatch \cite{berthelot2019mixmatch} for EEG learning, we first perform a data augmentation on \textit{labeled} data to obtain $\mathcal{A}_w(x_b^l)$. We then apply two different data augmentations on the \textit{unlabeled} data and calculate the model predictions:
\begin{equation}
q_b^{x}=p_m(y\mid \mathcal{A}_w(x_b^l);\theta) ~~~\text{and}
\end{equation} 
\begin{equation}
q_b^{u1}=p_m(y\mid \mathcal{A}_w(x_b^u);\theta), q_b^{u2}=p_m(y\mid \mathcal{A}_s(x_b^u);\theta),
\end{equation} 
where $q_b^{x}$ is the model prediction on augmented labeled inputs. $q_b^{u1}$ and $q_b^{u2}$ are the two model predictions on the same unlabeled input ($x_b^u$) with different augmentations.

In the second phase, we average the predictions over the augmented unlabeled data. We then minimize the entropy of the label distribution (a process known as sharpening), using:
\vspace{-3mm}
\begin{equation}\label{sharpen}
\small
\sharpen(p_b, T) = p_b^{1/T} \Big/\sum_1^k(p_b^{1/T}),
\end{equation} 
where $p_b$ is the average prediction as $p_b= (\softmax(q_b^{u1}) + \softmax(q_b^{u2}))/2$. $T$ is the `temperature' to adjust the entropy of the label distribution. We choose $T=1$ as suggested in \cite{berthelot2019mixmatch}. 

In the third phase, we use MixUp \cite{zhang2018mixup} to combine 
% the augmented labeled data with the label ($y^l$), 
% set 1: 
the augmented \textit{labeled} data along with the corresponding labels, with 
% and 
% set 2: 
the augmented \textit{unlabeled} data along with the sharpened predictions.
% as well as the augmented unlabeled data with the sharpened predictions, 
This creates new mixed labeled and unlabeled sets, ($\tilde{x}_b^l, \tilde{y}_b^l$) and ($\tilde{x}_b^u, \tilde{y}_b^u$) \cite{berthelot2019mixmatch}. The process is as follows:
\begin{equation}
\lambda \sim \mathcal{B}(\alpha, \alpha),
\end{equation} 
\begin{equation}
\lambda' = \max(\lambda, 1-\lambda),
\end{equation} 
\begin{equation}
\mathcal{M} = \{(\big< x_b^l, \mathcal{A}_w(x_b^u), \mathcal{A}_s(x_b^u)\big>, \big<y_b^l , p_b^u , p_b^u \big>)\},
\end{equation} 
\begin{equation}
\mathcal{M'} = \shuffle(\mathcal{M}),
\end{equation} 
\begin{equation}
(\tilde{x}_b^l, \tilde{y}_b^l) = \lambda' \{\mathcal{M}\}_{i=1}^{|x_b^l|}+ (1 - \lambda')\{\mathcal{M}'\}_{i=1}^{|x_b^l|},
\end{equation}
\begin{equation}
(\tilde{x}_b^u, \tilde{y}_b^u) = \lambda' \{\mathcal{M}\}_{i=|x_b^l|}^{|x_b^l|+2|x_b^u|}+ (1 - \lambda')\{\mathcal{M}'\}_{i=|x_b^l|}^{|x_b^l|+2|x_b^u|},
\end{equation}
where $\big< ,\big>$ represents the concatenation operation of two or more sets. $\mathcal{B}$ represents Beta distribution with $\alpha=0.75$. 
$p_b^{u}$ is the sharpened prediction based on $\sharpen(p_b, T)$. $\mathcal{M}$ is the concatenation of augmented labeled and unlabeled sets, $\mathcal{M}'$ is the set of $\mathcal{M}$ after a random shuffle operation, and $|.|$ is the size of the set. 

In the last step, we train the model on mixed data. Our semi-supervised loss comprises of a supervised term 
\begin{equation}\label{supervised_loss}
\mathcal{L}_s = \mathcal{H}(\tilde{y}_b^l, p_m(y\mid \tilde{x}_b^l;\theta)), 
\end{equation}
with cross-entropy loss ($\mathcal{H}$), and an unsupervised term 
\begin{equation}\label{unsupervised_loss}
\mathcal{L}_u = (p_m(y\mid \tilde{x}_b^u;\theta) - \tilde{y}_b^u)^2 .
\end{equation}
% with squared difference.
The overall semi-supervised loss is then calculated as $\mathcal{L}  = \mathcal{L}_s  +w \mathcal{L}_u$, where $w$ is a warm-up function
that is used to gradually increase the weight of the unsupervised term as the number of training epochs increases
\cite{berthelot2019mixmatch}.

\noindent \textbf{FixMatch.}
FixMatch, proposed in \cite{sohn2020fixmatch}, provides a simpler method for combining pseudo-labeling with consistency regularization. FixMatch employs cross-entropy between model predictions and pseudo-labels rather than calculating the squared difference as the unsupervised loss. The model prediction is computed using strongly augmented unlabeled data as $p_m(y\mid \mathcal{A}_s(x_b^u))$. The pseudo-label is generated using weakly augmented unlabeled data as $y_b^u= \argmax(q_b^u)$, where $q_b^u=p_m(y\mid \mathcal{A}_w(x_b^u);\theta)$. 

Following that, we utilize a pre-defined confidence threshold $\tau$ to guarantee that pseudo-labels are only used to update the loss when the model prediction is confident:
\begin{equation}\label{confidence}
\mathcal{L}_u = \ind(\max(q_b^u) \geq \tau)\cdot\mathcal{H}(y_b^u, p_m(y\mid \mathcal{A}_s(x_b^u);\theta)).
\end{equation} 
Here, $\ind$ is the binary mask and $\max(q_b^u)$ represents the highest class probability in the prediction ($q_b^u$).

Next, we use the supervised term \begin{equation}\label{supervised_fixmatch}
\mathcal{L}_s = \mathcal{H}(y_b^l, p_m(y\mid x_b^l;\theta)).
\end{equation} 
The total loss is the sum of the supervised and unsupervised terms as $\mathcal{L} = \mathcal{L}_u + \mathcal{L}_s$. No warm-up function is used since the weight of the unsupervised term will automatically rise as the model's confidence grows with each training epoch \cite{sohn2020fixmatch}.

\noindent \textbf{AdaMatch.}
To adapt AdaMatch \cite{berthelot2021adamatch}, which is an improved SSL method based on FixMatch, we first obtain the predictions for weakly augmented \textit{labeled} and \textit{unlabeled} data as $q_b^l=p_m(y\mid \mathcal{A}_w(x_b^l);\theta)$ and $q_b^u$, respectively, where $q_b^l,q_b^u \in \mathbb{R}^{B\times k}$. Let's represent the class distribution of \textit{labeled} data using the expected values of predictions for the weakly augmented \textit{labeled} data, as $\mathbb{E}{(q_b^l)} \in \mathbb{R}^k$, and the estimated class distribution of \textit{unlabeled} data using the expected values of predictions for the weakly augmented \textit{unlabeled} data, as $\mathbb{E}{(q_b^u)} \in \mathbb{R}^k$, where $\mathbb{E}(.)$ is expectation operator. Accordingly, the estimated labels for \textit{unlabeled} data are rectified by multiplying them with the ratio of class distributions of \textit{labeled} data to the estimated class distribution of \textit{unlabeled} data. This operation can be denoted as
\begin{equation}\label{rectification}
\tilde{q}_b^u=||q_b^u(\mathbb{E}{(q_b^l)}/\mathbb{E}{(q_b^u)})||,
\end{equation} where $\tilde{q}_b^u \in \mathbb{R}^{B\times k}$, and the L$1$ normalization operator $||.||$ ensures that the class distribution sums up to $1$. 

Following, the predefined threshold is multiplied by the mean value of the most confident prediction for the weakly augmented labeled data as $\tau \cdot \mu_{\max(q_b^l)} \in \mathbb{R}$, which is referred to as the relative confidence threshold \cite{berthelot2021adamatch}, where $\max(q_b^l) \in \mathbb{R}^B $. Consequently, the unsupervised loss is calculated as 
\begin{equation}\label{unsupervised_loss_adamatch}
% \begin{split} 
    \mathcal{L}_u = \ind(\max(\tilde{q}_b^u) \geq \tau \cdot \mu_{\max(q_b^l)}) \cdot \mathcal{H}(\tilde{y}_b^u, p_m(y\mid \mathcal{A}_s(x_b^u);\theta)),
% \end{split}
\end{equation} where $\tilde{y}_b^u = \argmax(\tilde{q}_b^u)$,  $\ind \in \{0,1\}^B$, $\max(\tilde{q}_b^u) \in \mathbb{R}^{B}$, and $\tilde{y}_b^u \in \{0,1 \}^{B \times k}$.

The total loss is the weighted sum of the supervised term ($\mathcal{L}_s$ in Eq. \ref{supervised_loss}) and the unsupervised term ($w \mathcal{L}_u$), where $w$ is a warm-up function used in \cite{berthelot2021adamatch}, as $\mathcal{L}  = \mathcal{L}_s  + w \mathcal{L}_u$.

\begin{table*}[t]
    \centering
    \scriptsize
    \setlength
    \tabcolsep{3.6pt}
    \caption{The performance of holistic approaches in comparison to other semi-supervised methods on SEED and SEED-IV.}
    \begin{tabular}{l|rrrrrr|rrrrrr}
    \toprule
    % \hline
        & \multicolumn{6}{c|}{SEED} & \multicolumn{6}{c}{SEED-IV} \\
        \cmidrule(l{3pt}r{3pt}){1-1} \cmidrule(l{3pt}r{3pt}){2-7}  \cmidrule(l{3pt}r{3pt}){8-13}
        Method / $m$                & 1 label         & 3 labels       & 5 labels       & 7 labels        & 10 labels      & 25 labels 
                                    & 1 label         & 3 labels       & 5 labels       & 7 labels        & 10 labels      & 25 labels      \\
        \cmidrule(l{3pt}r{3pt}){1-1} \cmidrule(l{3pt}r{3pt}){2-7}  \cmidrule(l{3pt}r{3pt}){8-13}
        $\Pi$-model	&	60.25 $ $\tiny(9.57)	&	67.87\tiny(10.14)	&	72.43\tiny(11.32)	&	74.94\tiny(10.84)	&	76.35\tiny(10.93)	&	77.87\tiny(10.88)
        	&	49.93\tiny(12.30)	&	54.65\tiny(14.66)	&	57.65\tiny(14.36)	&	58.79\tiny(14.77)	&	60.14\tiny(15.14)	&	61.92\tiny(15.30)	\\	
        
        Temporal Ens.	&	59.22 $ $\tiny(9.02)	&	69.95 $ $\tiny(9.07)	&	73.80 $ $\tiny(9.78)	&	77.15 $ $\tiny(9.57)	&	79.80 $ $\tiny(9.53)	&	83.83 $ $\tiny(8.73)
        	&	52.76\tiny(13.15)	&	59.76\tiny(14.48)	&	63.00\tiny(14.35)	&	65.26\tiny(14.07)	&	65.92\tiny(13.71)	&	67.25\tiny(13.63)	\\	
        
        Mean Teacher	&	53.97 $ $\tiny(8.24)	&	62.75 $ $\tiny(9.98)	&	66.42 $ $\tiny(9.46)	&	69.90\tiny(11.32)	&	71.48 $ $\tiny(8.98)	&	77.09 $ $\tiny(9.66)
        	&	47.03\tiny(11.84)	&	51.56\tiny(12.35)	&	55.05\tiny(13.27)	&	56.60\tiny(12.91)	&	56.66\tiny(11.78)	&	57.97\tiny(13.00)	\\	
        
        Conv. AutoEnc.	&	\underline{71.39\tiny(12.20)}	&	80.03\tiny(11.69)	&	82.86\tiny(10.89)	&	84.74 $ $\tiny(9.70)	&	85.46 $ $\tiny(9.77)	&	\underline{87.34 $ $\tiny(8.96)}
        	&	53.19\tiny(18.58)	&	59.52\tiny(18.13)	&	63.01\tiny(16.74)	&	64.83\tiny(15.75)	&	66.40\tiny(17.26)	&	65.96\tiny(16.62)	\\	
        
        Pseudo-Label	&	68.02\tiny(13.20)	&	78.11\tiny(12.02)	&	79.57\tiny(10.78)	&	82.21\tiny(11.03)	&	84.11 $ $\tiny(9.79)	&	85.32 $ $\tiny(9.38)
        	&	52.31\tiny(17.93)	&	58.08\tiny(16.76)	&	60.36\tiny(17.92)	&	60.84\tiny(17.59)	&	62.13\tiny(19.03)	&	62.71\tiny(18.36)	\\	
        
        \cmidrule(l{3pt}r{3pt}){1-1} \cmidrule(l{3pt}r{3pt}){2-7}  \cmidrule(l{3pt}r{3pt}){8-13}
        
        MixMatch	&	68.97\tiny(13.93)	&	\underline{80.89\tiny(12.80)}	&	\textbf{83.94\tiny(10.30)}	&	\underline{85.46 $ $\tiny(9.64)}	&	\underline{85.84 $ $\tiny(9.24)}	&	86.88 $ $\tiny(8.78)
        	&	\underline{56.08\tiny(15.92)}	&	\underline{65.03\tiny(15.79)}	&	\textbf{69.42\tiny(16.31)}	&	\textbf{70.92\tiny(16.02)}	&	\textbf{72.31\tiny(16.27)}	&	\textbf{73.20\tiny(15.19)}	\\	
        
        FixMatch	&	66.36\tiny(13.84)	&	76.26\tiny(11.56)	&	79.04\tiny(10.68)	&	81.79\tiny(10.56)	&	83.14 $ $\tiny(9.98)	&	84.44 $ $\tiny(9.09)
        	&	53.37\tiny(17.33)	&	63.57\tiny(15.57)	&	63.43\tiny(16.26)	&	64.62\tiny(15.57)	&	66.50\tiny(15.91)	&	68.54\tiny(15.58)	\\	
        
        AdaMatch	&	\textbf{74.03\tiny(11.78)}	&	\textbf{82.59\tiny(10.26)}	&	\underline{83.62\tiny(10.84)}	&	\textbf{85.84 $ $\tiny(9.69)}	&	\textbf{86.71 $ $\tiny(9.09)}	&	\textbf{88.02 $ $\tiny(8.80)}
        	&	\textbf{58.30\tiny(15.95)}	&	\textbf{66.52\tiny(16.58)}	&	\underline{69.12\tiny(16.45)}	&	\underline{68.11\tiny(15.80)}	&	\underline{69.31\tiny(16.87)}	&	\underline{71.43\tiny(16.04)}	\\	

        \bottomrule
    \end{tabular}
    \label{tab:realistic_ssl}
\end{table*}

\section{Experimental Setup and Results}
\subsection{Datasets}

\noindent \textbf{SEED.} The SEED dataset \cite{zheng2015investigating} contains EEG recorded using $62$ electrodes at the sampling rate of $1000$ \textit{Hz}. In the experiments, $15$ film clips with three emotions (neutral, positive, and negative) were selected as stimuli. The studies were completed by a total of $15$ individuals, $8$ females and $7$ males. Each subject takes part in the experiment twice, with each experiment consisting of $15$ recording sessions.

\noindent \textbf{SEED-IV.} The SEED-IV dataset \cite{zheng2018emotionmeter} contains $62$-channel EEG recordings obtained with a sample frequency of $1000$ \textit{Hz}. $72$ video snippets with four emotions (neutral, fear, sad, and happy) were used as stimuli. The experiments were carried out by $15$ participants ($8$ females and $7$ males). Each subject repeated the experiment three times, with different stimuli used each time. Each experiment has $24$ recording sessions ($6$ sessions for each emotion).

\vspace{-2mm}
\subsection{Pre-processing and Feature Extraction}
We follow the exact pre-processing steps described in \cite{zheng2015investigating,zheng2018emotionmeter} as follows. We first down-sample the EEG to $200$ \textit{Hz}. Then, to reduce artefacts, band-pass filters with frequencies ranging from $0.3-50$ \textit{Hz} and $1-75$ \textit{Hz} were applied to recordings from the SEED and SEED-IV. Following pre-processing, EEG were split into continuous segments of the same length ($1$ second for SEED and $4$ seconds for SEED-IV) with no overlap. Differential Entropy (DE) features were extracted from five bands (delta, theta, alpha, beta, and gamma) of each EEG segment. We assume EEG signals obey a Gaussian distribution $\mathcal{N}(\mu, \sigma^{2})$, and thus DE can be calculated as $DE = \frac{1}{2} \log{2\pi e \sigma^{2}}$.

\vspace{-3mm}
\subsection{Evaluation Protocols}
We apply the same evaluation protocols that were used in the original dataset publications \cite{zheng2015investigating,zheng2018emotionmeter}. For SEED, we use the first $9$ sessions for training and the remaining $6$ sessions for testing. For SEED-IV, we use the first 16 sessions for training, and the rest $8$ sessions for testing. We use the classification accuracy to measure the performance of the semi-supervised techniques for recognition of three emotions (neutral, positive, and negative) in SEED and four emotions (neutral, fear, sad, and happy) in SEED-IV.

\vspace{-2mm}
\subsection{Implementation details}

\noindent \textbf{Data Augmentation.}
We use additive Gaussian noise to augment the EEG signals. We denote the augmented data as $\mathcal{A}(x) = x + \mathcal{N}(\mu, \sigma)$, where $x\sim[0,1]$, $\mathcal{N}$ is a Gaussian distribution with $\mu=0.5$. The intensity level of data augmentation can be tuned by changing $\sigma$. We choose $\sigma$ of $0.8$ and $0.2$ in additive Gaussian Noise to represent strong ($\mathcal{A}_{s}$) and weak ($\mathcal{A}_{w}$) augmentations, respectively.

\noindent \textbf{Backbone Network.}
As our backbone network, we use an efficient lightweight CNN architecture, consisting of two convolutional blocks and one classifier. Each block consists of a $1$-D convolutional layer followed by a $1$-D batch normalization layer and a LeakyReLU activation. The classifier contains two fully connected layers with a dropout rate of $0.5$.

\noindent \textbf{Hyper-parameters.}
In all the experiments, we run $30$ training epochs with a batch size of $8$. We employ Adam optimizer \cite{kingma2014adam} with the default learning rate of $0.001$. Tuning of the hyper-parameters  has been done on the validation set $D_v$ (with no leakage with the test set). To set the pre-defined threshold ($\tau$) in FixMatch and AdaMatch, we perform a grid search in $[0.0-1.0]$ with a step size of $0.1$. In FixMatch, we set $\tau = 0.9$ for both datasets, while in AdaMatch, we set $\tau = 0.6$ and $\tau = 0.5$ for SEED and SEED-IV, respectively. Our experiments were carried out on two NVIDIA GeForce RTX $2080$ Ti and six Tesla P$100$ GPUs using PyTorch \cite{paszke2019pytorch}.

\vspace{-3mm}
\subsection{Classical SSL Settings}
For all the classical SSL baselines, we use the same convolutional module used as the backbone for the holistic methods (see Section 3.4). For the decoder component of the convolutional autoencoder, we use two transposed convolutional $1$-D blocks. In each block, a $1$-D transposed convolutional layer is followed by a $1$-D batch normalization layer and a ReLU activation. We implement the $\Pi$-model, temporal ensembling, mean teacher, pseudo-labeling, and convolutional autoencoder with the same algorithm settings (e.g., loss function, unsupervised weight, etc.) as used in \cite{zhang2021deep}.

\begin{figure}[t!] 
    \begin{center}
    \includegraphics[width=0.80\columnwidth]{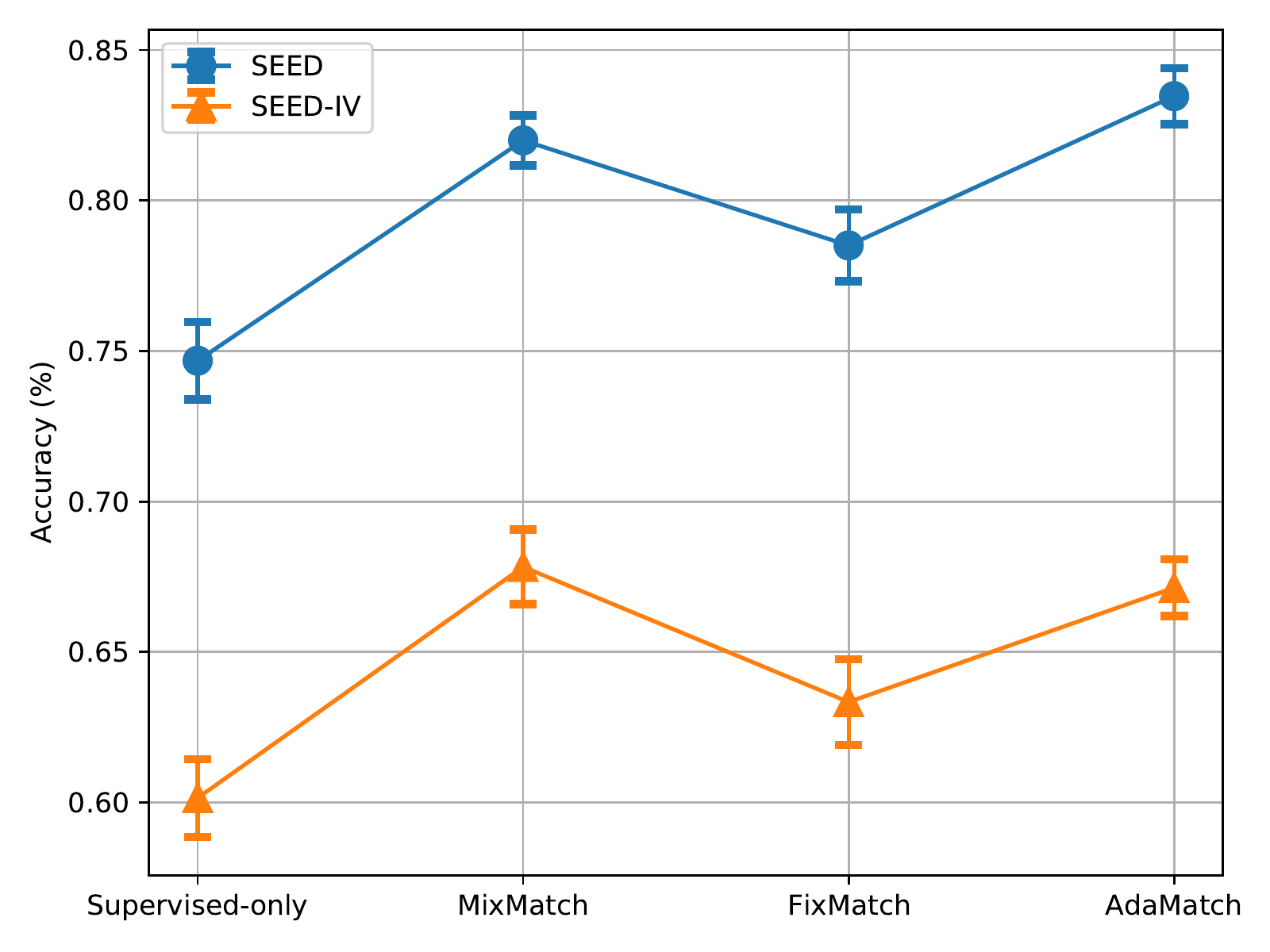} 
    \end{center}
\vspace{-7mm}
\caption{Average performance of holistic SSL methods in comparison to a supervised method.}
\vspace{-5mm}
\label{fig: barplot}
\end{figure}

\vspace{-3mm}
\subsection{Performance}
We conduct experiments using all the SSL methods where a few training samples per class are labeled ($m \in \{1, 3, 5, 7, 10,$ $25 \}$) , as done in \cite{berthelot2019mixmatch,sohn2020fixmatch,berthelot2021adamatch}. We repeat each experiment \textbf{five} times, each time using a different random seed for the selection of the subsets ($D_l, D_u$, and $D_v$). Table \ref{tab:realistic_ssl} displays the average and standard deviations of accuracies for SEED and SEED-IV datasets. Our results show that in the majority scenarios where very few labeled samples are used, the holistic SSL approaches not only achieve the best results (shown in bold), but also obtain the second-best performances (shown with underline). In SEED, AdaMatch achieves the best performance, while MixMatch achieves the second-best results. AdaMatch performs the best in SEED-IV when very few labeled samples are available ($m \in \{1, 3\}$), while MixMatch performs the best when more samples are labeled ($m \in \{5, 7, 10, 25\}$). AdaMatch and MixMatch consistently rank first and second in most scenarios for both datasets. Among the classical SSL methods, the Convolutional Autoencoder consistently achieves the best results for both datasets.

We compare the performance of the holistic approaches to the supervised-only method, where the results are averaged across all six scenarios where few labeled samples are present. The supervised-only method employs the same backbone network (Section 3.4) but are \textit{only} trained on the labeled samples. As shown in Figure \ref{fig: barplot}, AdaMatch achieves the best performance with $83.47 \pm 10.07\%$ for SEED, while MixMatch achieves the best result with $67.83 \pm 15.92\%$ for SEED-IV. AdaMatch and MixMatch outperform FixMatch in both datasets. In comparison, the supervised-only model achieves accuracies of $74.68 \pm 11.32\%$ and $60.15 \pm 16.28\%$ for SEED and SEED-IV, respectively.

\vspace{-3mm}

\section{Conclusion} 
In this research, we adapt and implement three state-of-the-art holistic semi-supervised learning approaches originally proposed for computer vision \cite{berthelot2019mixmatch,sohn2020fixmatch,berthelot2021adamatch}, for EEG representation learning. We conduct extensive experiments with the holistic SSL approaches and five additional well-known classical SSL methods, where only $1,3,5,7,10,$ and $25$ samples per class are labeled. We evaluate all the SSL methods on two large publicly available datasets, namely SEED and SEED-IV. The holistic approaches achieve the best results and the second-best results in the majority of scenarios for both datasets. The results also show that the holistic approaches remarkably outperform the supervised-only method, addressing the challenge of scarcity of labeled EEG data.

\bibliographystyle{IEEEbib}
\bibliography{refs}

\begin{thebibliography}{10}

\bibitem{berthelot2019mixmatch}
David Berthelot, Nicholas Carlini, Ian Goodfellow, Nicolas Papernot, Avital
  Oliver, and Colin~A Raffel,
\newblock ``Mixmatch: A holistic approach to semi-supervised learning,''
\newblock {\em Advances in Neural Information Processing Systems (NeurIPS)},
  vol. 32, 2019.

\bibitem{sohn2020fixmatch}
Kihyuk Sohn, David Berthelot, Nicholas Carlini, Zizhao Zhang, Han Zhang,
  Colin~A Raffel, Ekin~Dogus Cubuk, Alexey Kurakin, and Chun-Liang Li,
\newblock ``Fixmatch: Simplifying semi-supervised learning with consistency and
  confidence,''
\newblock {\em Advances in Neural Information Processing Systems (NeurIPS)},
  vol. 33, 2020.

\bibitem{berthelot2021adamatch}
David Berthelot, Rebecca Roelofs, Kihyuk Sohn, Nicholas Carlini, and Alex
  Kurakin,
\newblock ``Adamatch: A unified approach to semi-supervised learning and domain
  adaptation,''
\newblock {\em arXiv preprint arXiv:2106.04732}, 2021.

\bibitem{picard2000affective}
Rosalind~W Picard,
\newblock {\em Affective Computing},
\newblock MIT press, 2000.

\bibitem{zheng2015investigating}
Wei-Long Zheng and Bao-Liang Lu,
\newblock ``Investigating critical frequency bands and channels for eeg-based
  emotion recognition with deep neural networks,''
\newblock {\em IEEE Transactions on Autonomous Mental Development}, vol. 7, no.
  3, pp. 162--175, 2015.

\bibitem{zheng2018emotionmeter}
Wei-Long Zheng, Wei Liu, Yifei Lu, Bao-Liang Lu, and Andrzej Cichocki,
\newblock ``Emotionmeter: A multimodal framework for recognizing human
  emotions,''
\newblock {\em IEEE Transactions on Cybernetics}, vol. 49, no. 3, pp.
  1110--1122, 2018.

\bibitem{zhang2021capsule}
Guangyi Zhang and Ali Etemad,
\newblock ``Capsule attention for multimodal eeg-eog representation learning
  with application to driver vigilance estimation,''
\newblock {\em IEEE Transactions on Neural Systems and Rehabilitation
  Engineering}, 2021.

\bibitem{zhang2020rfnet}
Guangyi Zhang and Ali Etemad,
\newblock ``Rfnet: Riemannian fusion network for eeg-based brain-computer
  interfaces,''
\newblock {\em arXiv preprint arXiv:2008.08633}, 2020.

\bibitem{zheng2017identifying}
Wei-Long Zheng, Jia-Yi Zhu, and Bao-Liang Lu,
\newblock ``Identifying stable patterns over time for emotion recognition from
  eeg,''
\newblock {\em IEEE Transactions on Affective Computing}, 2017.

\bibitem{zhang2018spatial}
Tong Zhang, Wenming Zheng, Zhen Cui, Yuan Zong, and Yang Li,
\newblock ``Spatial-temporal recurrent neural network for emotion
  recognition,''
\newblock {\em IEEE Transactions on Cybernetics}, , no. 99, pp. 1--9, 2018.

\bibitem{correa2018amigos}
Juan Abdon~Miranda Correa, Mojtaba~Khomami Abadi, Niculae Sebe, and Ioannis
  Patras,
\newblock ``Amigos: A dataset for affect, personality and mood research on
  individuals and groups,''
\newblock {\em IEEE Transactions on Affective Computing}, 2018.

\bibitem{van2020survey}
Jesper~E Van~Engelen and Holger~H Hoos,
\newblock ``A survey on semi-supervised learning,''
\newblock {\em Machine Learning}, vol. 109, no. 2, pp. 373--440, 2020.

\bibitem{lee2013pseudo}
Dong-Hyun Lee et~al.,
\newblock ``Pseudo-label: The simple and efficient semi-supervised learning
  method for deep neural networks,''
\newblock in {\em Workshop on challenges in representation learning, ICML},
  2013, vol.~3, p. 896.

\bibitem{samuli2017temporal}
Laine Samuli and Aila Timo,
\newblock ``Temporal ensembling for semi-supervised learning,''
\newblock in {\em International Conference on Learning Representations (ICLR)},
  2017, vol.~4, p.~6.

\bibitem{tarvainen2017mean}
Antti Tarvainen and Harri Valpola,
\newblock ``Mean teachers are better role models: Weight-averaged consistency
  targets improve semi-supervised deep learning results,''
\newblock {\em Advances in Neural Information Processing Systems (NeurIPS)},
  vol. 30, 2017.

\bibitem{oliver2018realistic}
Avital Oliver, Augustus Odena, Colin~A Raffel, Ekin~Dogus Cubuk, and Ian
  Goodfellow,
\newblock ``Realistic evaluation of deep semi-supervised learning algorithms,''
\newblock {\em Advances in Neural Information Processing Systems (NeurIPS)},
  vol. 31, pp. 3235--3246, 2018.

\bibitem{zhang2021deep}
Guangyi Zhang and Ali Etemad,
\newblock ``Deep recurrent semi-supervised eeg representation learning for
  emotion recognition,''
\newblock in {\em 2021 9th International Conference on Affective Computing and
  Intelligent Interaction (ACII)}. IEEE, 2021, pp. 1--8.

\bibitem{tong2019caesnet}
Li~Tong, Hang Wu, and May~D Wang,
\newblock ``Caesnet: Convolutional autoencoder based semi-supervised network
  for improving multiclass classification of endomicroscopic images,''
\newblock {\em Journal of the American Medical Informatics Association}, vol.
  26, no. 11, pp. 1286--1296, 2019.

\bibitem{zhang2018mixup}
Hongyi Zhang, Moustapha Cisse, Yann~N Dauphin, and David Lopez-Paz,
\newblock ``mixup: Beyond empirical risk minimization,''
\newblock in {\em International Conference on Learning Representations (ICML)},
  2018.

\bibitem{kingma2014adam}
Diederik~P Kingma and Jimmy Ba,
\newblock ``Adam: A method for stochastic optimization,''
\newblock {\em arXiv preprint arXiv:1412.6980}, 2014.

\bibitem{paszke2019pytorch}
Adam Paszke, Sam Gross, Francisco Massa, Adam Lerer, James Bradbury, Gregory
  Chanan, Trevor Killeen, Zeming Lin, Natalia Gimelshein, Luca Antiga, et~al.,
\newblock ``Pytorch: An imperative style, high-performance deep learning
  library,''
\newblock {\em Advances in Neural Information Processing Systems (NeurIPS)},
  vol. 32, pp. 8026--8037, 2019.

\end{thebibliography}

\end{document}